\documentclass[runningheads]{llncs}
\usepackage{graphicx}
\usepackage{comment}
\usepackage{amsmath,amssymb}
\usepackage{color}
\usepackage{cite}
\usepackage{caption}

\usepackage{nccmath}

\usepackage{amssymb}
\usepackage{booktabs}
\usepackage{multirow}
\usepackage{float}
\usepackage[font=footnotesize,labelfont=bf]{caption}

\newcommand{\cmmnt}[1]{}

\begin{document}
\pagestyle{headings}
\mainmatter
\title{A Few-Shot Sequential Approach for Object Counting} 

\titlerunning{A Few-Shot Sequential Approach for Object Counting} 
\authorrunning{N. Sokhandan, P. Kamousi, A. Posada, E. Alese, N. Rostamzadeh}

\author{Negin Sokhandan \hspace{5mm} Pegah Kamousi \hspace{5mm} Alejandro Posada  \hspace{5mm} \\ Eniola Alese \hspace{5mm} Negar Rostamzadeh}
\institute{Element AI \\
{\tt\small \{negin, pegah, alejandro.posada, eniola.alese, negar\}@elementai.com}}

\maketitle
\begin{abstract}
In this work, we address the problem of few-shot multi-class object counting with point-level annotations.
The proposed technique leverages a class agnostic attention mechanism that sequentially attends to objects in the image and extracts their relevant features. This process is employed on an adapted prototypical-based few-shot approach that uses the extracted features to classify each one either as one of the classes present in the support set images or as background.
The proposed technique is trained on point-level annotations and uses a novel loss function that disentangles class-dependent and class-agnostic aspects of the model to help with the task of few-shot object counting. We present our results on a variety of object-counting/detection datasets, including FSOD and MS COCO. In addition, we introduce a new dataset that is specifically designed for weakly supervised multi-class object counting/detection and contains considerably different classes and distribution of number of classes/instances per image compared to the existing datasets. We demonstrate the robustness of our approach by testing our system on a totally different distribution of classes from what it has been trained on.
\end{abstract}

\section{Introduction}
Object counting is an important task in computer vision motivated by a variety of applications such as traffic monitoring, wildlife conservation and retail inventory tracking. Several methods focus on counting objects of a single class, such as people~\cite{li2018csrnet, sindagi2017cnn, zhang2015cross, zhang2016single}, cars~\cite{de2015pklot, onoro2016towards} or cells~\cite{khan2016deep, paul2017count, xie2018microscopy}. However, multi-class counting methods are more relevant to real-world applications such as counting items on supermarket shelves with several items of multiple categories in an image.

While deep learning techniques have revolutionized the field of computer vision in the past decade, the performance of such models often comes at the cost of acquiring large amounts of labelled data. This poses a great challenge for object counting in particular, where acquiring per-item labels with possibly many items per image increases the cost of labelling dramatically. This motivates the development of training strategies that enable the models to recognize and count new categories given only a few labeled images. 

\begin{figure}[t]
\center
    \hspace*{1.4cm}
    \includegraphics[width=0.58\textwidth]{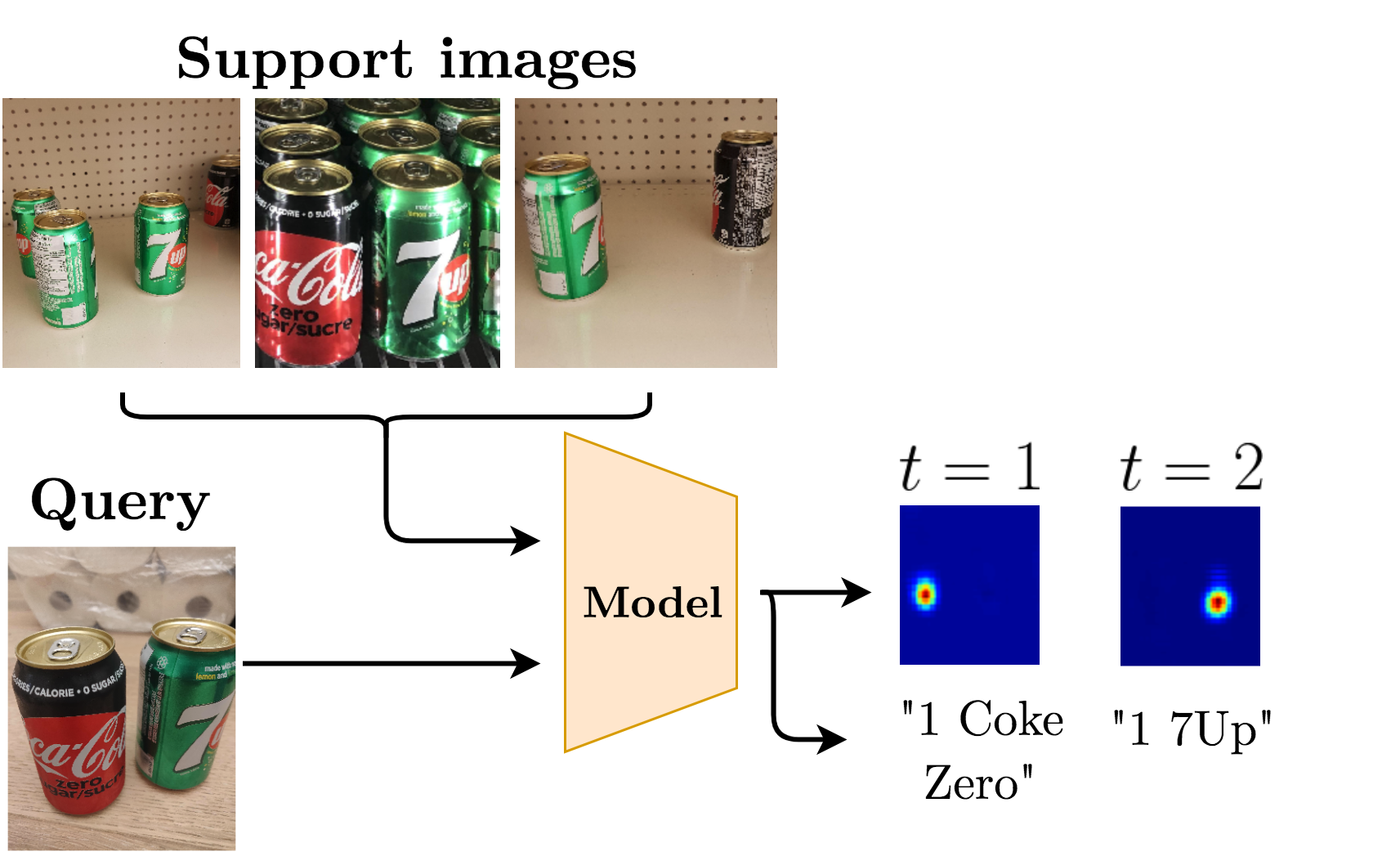}
    \caption{Given support and query images of the same task, the model sequentially classifies and pays attention to each object in the image.}
    \label{fig: task}
\end{figure}

Unlike most existing deep learning approaches, humans are capable of learning to count new objects from unseen categories relying only on a few examples. Few-shot learning attempts to enable such data efficiency in machine perception with the goal of training models in low-data regimes where few labelled examples are available for each task. Most existing few-shot learning methods can be categorized into two groups: \text{1) gradient-based methods} \cite{andrychowicz2016learning, finn2017model, nichol2018first}, which rely on a \textit{meta-learner} that predicts the parameters of task-specific models, and \text{2) metric-based methods} ~\cite{koch2015siamese, snell2017prototypical, vinyals2016matching}, which learn a similarity measure to compare a query image against a labelled support set. However, most methods focus on image classification and adapting them to the more complex task of multi-class object counting is nontrivial: object counting requires to both localize and classify each item instead of merely relying on a global comprehension of the image. 

In this work, we address the problem of few-shot multi-class object counting using only point-level annotation, wherein only one pixel from each object is annotated. Such setup mitigates the challenge of acquiring large amounts of labelled data by reducing not only the number of required training samples but also the cost of labelling each sample.

Experiments on visual cognition in humans suggest that we do not tend to focus our attention on the entire scene at once. Instead, we attend sequentially to different parts in order to extract relevant information~\cite{rensink2000dynamic}. This appears to be particularly effective in higher level cognitive tasks such as counting objects of multiple categories in a scene, where we tend to zoom into one object at a time~\cite{wilder2009attention}. Similarly, our model uses an attention mechanism that sequentially extracts features of the objects in a query image in a specific order. Those features are then compared to the class prototypes extracted from the support images in order to be classified. We focus on scenarios with average density and large number of classes per image, since unlike high-density, single-class tasks such as crowd counting \cite{ sindagi2017cnn ,zhang2015cross, zhang2016single}, this setting has rarely been addressed in the existing literature. Our contributions can be summarised as follows.

\begin{enumerate}
\item We propose a novel recurrent attention-based system that sequentially computes one attention map per each object in the query image. The maps are then used to weight the feature vectors, which are in turn used to classify each object by comparing it against a set of prototype extracted from the support images (Figure \ref{fig: task}). The labels are sorted in lexicographical order by their $(y, x)$ coordinates, guiding the model to attend to the objects in the query images in the same order.
\item We use a novel loss function that consists of a class-agnostic and a class-dependent term. The former helps fit the attention map at each time-step to a Gaussian distribution, hence localizing objects in the image, while the latter encourages the model to classify those items correctly. The ratio of these two losses changes throughout the training, assigning a larger weight on the class-agnostic term at first and exponentially decaying it as the training proceeds.
\item We introduce a dataset with a distribution of classes and objects that is considerably different from images of natural scenes in publicly available datasets. The objects in our dataset come from various categories of grocery items such as sodas, canned food, etc. The average density of objects and the variety of classes \textit{in each image} makes the dataset suitable for evaluating and benchmarking few-shot techniques.
\end{enumerate}

\section{Related Work}
Given that our contributions can be stated both in terms of our approach to multi-class object counting as well as to few-shot learning, in this section we briefly discuss the existing approaches in both domains.

\subsection{Object counting}
Object counting methods can be roughly divided into two categories: detection-based and regression-based.

\textbf{Regression-based methods} rely on regressors to estimate the object counts. A variety of successful approaches from heuristic-based to deep learning methods belong to this category ~\cite{chan2008privacy,  chan2009bayesian, chen2012feature, cholakkal2019object, kong2005counting, liu2015bayesian, liu2019point, marana1998efficacy, ryan2009crowd, sam2019locate, ShiICCV19, zou2019attend}, with
Glance~\cite{chattopadhyay2017counting} and density-based methods being among the most successful examples.  
Glance~\cite{chattopadhyay2017counting} uses image-level labels, \textit{i.e.} per-class counts, and learns to estimate the global count in a single forward pass. Glance is efficient with small counts. However, it employs ``subitizing'' technique for large counts, which is hard to train and requires bounding box labels. Density-based approaches learn to count by regressing a density map using a least-squares objective and obtain the total count by integrating over the density maps~\cite{boominathan2016crowdnet, cheng2019learning, li2018csrnet, marsden2016fully, onoro2016towards, sindagi2017cnn, sindagi2019pushing, walach2016learning, yan2019perspective, zhang2016single}. Some of the density-based methods use a multi-column-based architecture~\cite{onoro2016towards, shang2016end, zhang2016single}, which introduces redundant structures, while others ~\cite{sindagi2017cnn} use a high-level prior to guide the computation of the density maps. These approaches often assume fixed object sizes defined by Gaussian kernels or constrained environments, which work well on crowd-counting problems but not with objects of varying sizes.

\textbf{Detection-based methods} can generalize better for objects with different sizes. These methods first detect the objects and then count the number of detected instances. Most such approaches~\cite{liu2016ssd, ren2015faster} rely on bounding box labels, which are not only expensive to acquire but also make the model prone to occlusions for densely packed images. To overcome this issue~\cite{trax} proposes a soft-IoU layer to estimate the overlap between predicted and ground truth bounding boxes, along with an expectation maximization\cmmnt{, i.e, EM-based} unit that clusters the Gaussians into groups to resolve overlap ambiguities. However, employing the EM-based step slows the model down significantly while relying on bounding-box annotation makes labelling costly.~\cite{laradji2018blobs} propose employing only point-level annotations to train a model that outputs one blob per instance. This approach mitigates the problem of occlusion in detection-based methods by not relying on bounding boxes. From the object localization perspective, our approach is closely related to the detection based approaches and following~\cite{laradji2018blobs} we rely only on point-level annotation for our counting model, which are relatively cheap to acquire. \newline

\cmmnt{~\cite{liu2019point, sam2019locate} train dense detectors on pseudo box labels built from point labels. ~\cite{ShiICCV19} also re-purpose the point annotations by building binary maps to focus on areas of interest. ~\cite{cholakkal2019object} use image-level supervision to produce a density map of the objects. ~\cite{zou2019attend} leverage an attention mechanism and three scale-specific networks to estimate the crowd density.}

\vspace{-5mm}
\subsection{Few-shot learning}
Few-shot learning aims to train models that generalize to new tasks using only a few samples, leveraging prior knowledge. Some early methods follow a Bayesian framework that learns to incorporate a prior such as saliency~\cite{fei2006one} or strokes and object parts ~\cite{lake2011one, lake2015human, lake2013one}. Image hallucination is used in~\cite{hariharan2017low, wang2018low, zhang2018metagan} to augment the training data to better generalize to new classes. Broadly speaking, there are two main categories of few-shot learning approaches (i) gradient-based and (ii) metric-based approaches.

\textbf{Gradient-based methods} aim at training models that generalize well to new tasks/categories with only a few fine-tuning updates ~\cite{ravi2016optimization}. The Model-agnostic meta-learning approach, MAML~\cite{finn2017model}, learns to adapt the weights to new task in a few gradient steps.  Many recent approaches have built upon the success of MAML~\cite{nichol2018reptile, lacoste2018uncertainty, jiang2018learning, nichol2018first, metasgd, bmaml2018, finnXL18, rusu2018meta}.
These approaches require fine-tuning and additional optimization steps. In contrast, our model addresses unseen tasks in a feed-forward manner following metric-based approaches, thus avoiding further gradient computations and model updates, which could create additional complication for complex tasks such as object counting.
\cmmnt{Due to the success of MAML, many recent gradient-based approaches built to improve its performance~\cite{nichol2018reptile},~\cite{lacoste2017deep},~\cite{jiang2018learning}. Reptile~\cite{nichol2018first} is a first-order approximation of MAML while Meta-SGD~\cite{metasgd} learns a set of parameters to control the gradient steps. BMAML \cite{bmaml2018} and ~\cite{finnXL18} propose a probabilistic extension to MAML with a variational approximation. Latent embedding optimization, LEO~\cite{rusu2018meta} address MAML's problem of only using a few updates on a low data regime to train models in a high dimensional parameter space. This elegant approach achieves current state of the art result in different few-shot classification benchmarks. 
Other meta-learning approaches for few-shot learning include using memory architecture to either store exemplar training samples~\citep{santoro16mem} or to  directly encode fast adaptation algorithm~\citep{ravi2016optimization}. Mishra et al. ~\citep{mishra2018simple} use temporal convolution to achieve the same goal.
These methods need to fine-tune and perform additional optimization steps. In contrast, our model addresses unseen tasks in a feed-forward manner, thus avoiding further gradient computations and model updates, that is beneficial for more complicated tasks such as object counting.}

\textbf{Metric-based methods} learn a distance metric to compare query images against the support images.~\cite{koch2015siamese} uses a siamese network to capture the similarity between images.~\cite{snell2017prototypical, vinyals2016matching} propose a matching network that learns a differentiable $k$-nearest neighbor model. ~\cite{sung2018learning} present a relation network that learns the optimal distance metric.~\cite{kim2019edge, bruna} use graph neural networks to model the relationship between support and query images. Due to the simplicity and adaptability of metric-based approaches, we base our approach on this group of work and in particular prototypical networks ~\cite{snell2017prototypical}.

\textbf{Few-shot counting, detection and segmentation:} the majority of approaches in few-shot learning focus on the problem of object classification. A few recent approaches have been devoted to address problems such as few-shot object detection~\cite{chen2018lstd, kang2019few} and segmentation ~\cite{dong2018few, HuAAAI2019, wang2019panet}. One of the approaches most relevant to ours is~\cite{rakelly2018few}, where given sparse point annotations they extract support and query features, which are in turn fed to a decoder that generates segmentation results. 
Even though not framed as a few-shot counting approach,~\cite{lu2018class} proposes a class-agnostic counting approach using a matching network which takes as input a query image and an exemplar patch containing the object of interest. The outputs of the network are then fed into a discriminative classifier. Finally, to adapt the network to novel categories, a small fraction of the learned parameters are fine-tuned using a few labelled examples. To the best of our knowledge, no prior work has been done specifically on few-shot object counting. In the next sections we define this problem and our proposed approach.  

\section{Problem Definition} \label{problem_def}
We follow the setup common in few-shot classification and segmentation tasks ~\cite{rakelly2018few, snell2017prototypical}, where meta-test classes are disjoint from the meta-train classes. We use episodic training, where the input data is divided into mini-batches or \textit{episodes}. In each episode, the input data is divided into a support set of $S$ annotated images or \textit{shots} used for supervision and a query set of $Q$ images to perform the task on. All the support and query images within an episode share the same \textit{task}, namely the same subset of $C$ classes from the $N$ total classes in the dataset.\cmmnt{The loss in each episode is optimized for the query images given the support shots of the episode.} In this paper, we use training and testing to refer to what the model does within each given episode, while \textit{meta-training} and \textit{meta-testing} refer to the process of teaching the model to adapt to new tasks. To summarize, the training proceeds in episodes, where each episode is a mini-batch of meta-train samples all taken from the same task.

We pose the problem as \textit{weakly-supervised few-shot object counting}, where only point-level annotations are available, \textit{i.e.} for each object in the image there is only a single annotated pixel anywhere on the area of the object, with a label indicating its class. In addition, we assume that multiple instances from each of the $C$ classes may be present in each image for both support and query images. We also assume the number of instances per class varies by class and image.
Following the notation in~\cite{rakelly2018few}, a few-shot object counting task with point-level supervision is defined as a set of input-output pairs $(\mathcal{T}_i, \mathcal{Y}_i)$ sampled from a task distribution $P$. The task inputs are 
\begin{align}
    \mathcal{T} &= \{\{(x_1,L_1), \dots, (x_S, L_S)\}, \{\bar{x}_1,    \dots,\bar{x}_Q\}\} \\
    \begin{split}
    L_s &= \{(p_i, l_i): i \in \{1, \dots, n_s\} \}, \\
    l &\in \{1, \dots, C\}
    \cup \{\varnothing\},
    \end{split}
\end{align}
where $x_s$ are the support images, $\bar{x}_q$ the query images, $L_s$ the annotation set for the $s$-th support image, $p_i$ and $l_i$ the the point and class labels for the $i$-th object in $x_s$, and $n_s$ the number of objects in $x_s$. Finally, $\varnothing$ is the background class. The target outputs are
\begin{align}
    \mathcal{Y} &= \{y_1, \dots, y_Q\}, y_q = \{(p_j, l_j) : j \in \{1, \dots, n_q\} \}
\end{align}
where $n_q$ is the number of objects in the $q$-th query image.

\begin{figure*}[t]
    \centering
    \includegraphics[width=1\textwidth]{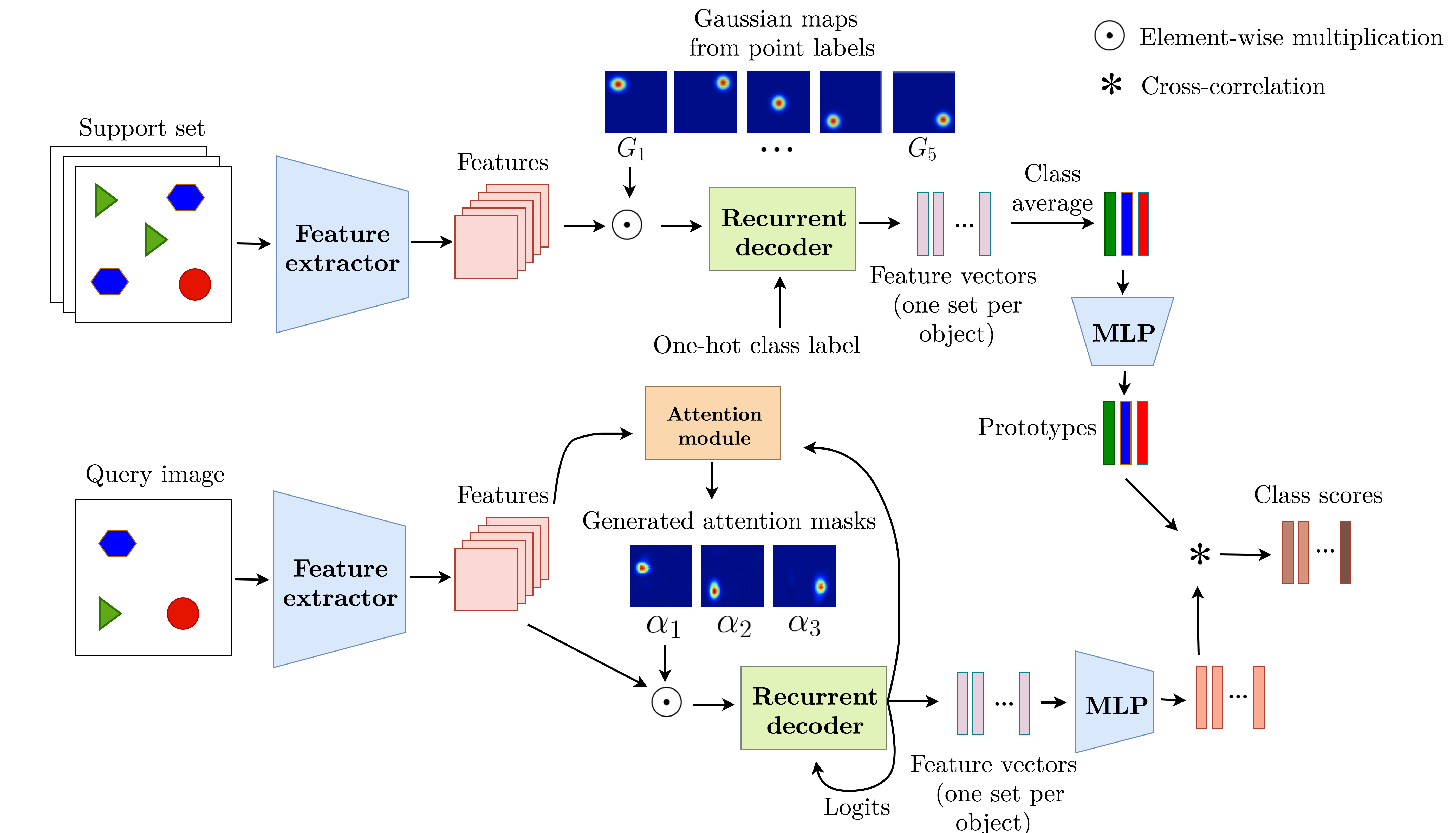}
    \caption{Architecture of our model. The support set images and the query image are fed to a shared feature extractor. The feature maps from each image are passed through the decoder. The decoder uses a sequential attention mechanism to firstly generate a weighting map for each object in the image and then use the maps to generate a feature vector for each object. For the support images instead of using the attention module to generate the feature maps, the Gaussian maps generated from the label points are used as weighting maps to create prototype feature vectors for each class. Class scores are computed by cross-correlating the query feature vectors with prototype feature vectors.}
    \label{fig: model}
\end{figure*}

\section{Proposed Method}
We approach the problem of object counting from a rather intuitive point of view that fits the problem nicely into a few-shot framework. We approach object counting as a two-step process of firstly localizing and distinguishing the objects from the background and secondly classifying each of them as one of the classes in the support set. The localization step is fundamentally independent of the class labels and can therefore be learned from the massive publicly available datasets. This allows the model to be as general as possible and mitigates the data-scarcity problem. The second step, on the other hand, is specific to the classes of interest and has to be learned from the few support images. To explain this further, object counting can be thought of as an image captioning task (similar to  \cite{Xu2015showattend}) wherein the model describes what is present in the image by paying attention to all of the objects in the image sequentially and in a specific predetermined order, outputting the classes of the objects that it attends to as a description for the image. The architecture of our proposed system is illustrated in Figure \ref{fig: model}. Our model consists of a fully-convolutional feature extractor and an attention-based recurrent decoder that sequentially outputs the class of the object to which it is attending. We explain in detail the components of the architecture in the following subsections.
\subsection{Feature Extractor}
We use ResNet-50~\cite{he2016deep} up to the fully-connected layers as the backbone of our feature extractor module. In order to improve the recognition of objects with different scales, we concatenate the features from four different layers of the backbone network (\texttt{conv2\_x}, \texttt{conv3\_x}, \texttt{conv4\_x} and \texttt{conv5\_x} explained  in ~\cite{he2016deep}) after up-sampling them to a fixed size, namely, $1/4$ of the original image. To make the model more location-aware, we concatenate the encoded location of each pixel in the feature map to the features extracted from the image. This is done by concatenating the one-hot encoding of the $x$ and $y$ coordinates together. The constructed feature maps $f$ are then processed by a decoder that is explained in the next subsection. 

\subsection{Decoder}\label{decoder}
\begin{figure}[t]
    \centering
    \includegraphics[width=.8\textwidth]{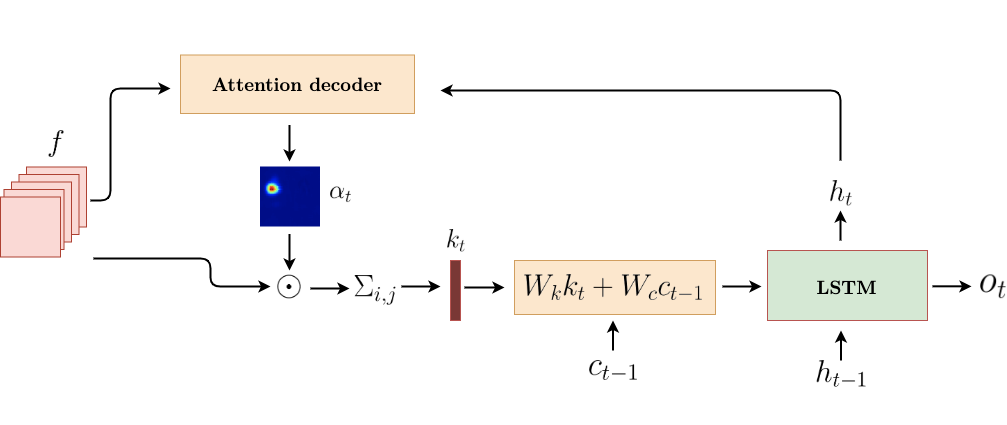}
    \caption{Architecture of the decoder. The extracted image features and the previous LSTM state are used to generate an attention map $\alpha_t$ (Equation \ref{eq: alpha}). The generated map is used to weigh the feature maps and compute a feature vector for the current time-step, $k_t$, which is then linearly combined with the predicted class scores at the previous time-step to form the input of the LSTM. The output of the LSTM is then used to compute the current class scores using the class prototypes.}
    \label{fig: decoder}
\end{figure}

The decoder component of this model, similar to the one proposed in~\cite{wojna2017attention}, uses the extracted features to sequentially output a class index for each of the objects present in the image. This is done by combining an RNN (specifically an LSTM ~\cite{hochreiter1997long}) and an attention module that interact together in a loop (Figure \ref{fig: decoder}). At each time-step $t$, the attention module generates an attention map $\alpha_t$ by linearly combining the image features $f$ and the previous LSTM state $h_{t-1}$ and passing the output through a non-linearity as follows:
\begin{align} \label{eq: alpha}
    \alpha_t &= \text{softmax}(v^\intercal \text{tanh}(W_h h_{t-1} + W_{f} f)),
\end{align}
where $W_h$, $W_f$ and $v$ are trainable weights. The generated attention maps are then used to spatially weigh the image feature maps and thus reduce them into a feature vector, $k_t$:
\begin{align}  \label{eq: weighting}
    k_{t} = \sum_{i,j} \alpha_{t,i,j} f_{i,j}.
\end{align}

This feature vector is then linearly combined with the predicted class scores vector from the previous time step, $c_{t-1}$, to form the next input to the LSTM:
\begin{align} \label{eq: xt}
    x_t &= W_k k_{t} + W_{c} c_{t-1},
\end{align}
where $W_k$ and $W_c$ are trainable weights. Finally, the output of the LSTM is used to generate class scores for the current time-step:
\begin{align} \label{eq: lstm}
    o_t, h_t = \text{LSTMstep}(x_t, h_{t-1}) ,\;\; 
     c_t = \text{softmax}(o_t).
\end{align}
When computing the class prototype features from the support set, instead of using the attention module to generate the weight maps, we use the annotation point of the current object and an estimate of the standard deviation $\sigma$ (as explained in \cite{zhang2016single}) to generate a Gaussian kernel $G_{t}^s$. This kernel is centered on the object of interest and is used to weigh the features extracted from the support image. We also generate a weight map $G_{\varnothing}^s$ for the background class:
\begin{align} \label{eq: xt}
    G_{\varnothing}^s &= \textbf{1}_{M \times N} - \sum^{n_s} G_t^s,
\end{align}
where $\textbf{1}_{M \times N}$ is an all-ones matrix with same dimensions as the maps. 
The feature vectors generated from the support images are averaged over all of the objects from the same class in the support set. The resulting features are used as class prototypes to classify the objects extracted from the query image. In order to construct class prediction logits for the query objects, the extracted feature vector for each object in the query is cross-correlated to all of the class prototypes from the support set, including the background. All of the query feature vectors and prototype feature vectors are passed through a linear layer prior to cross-correlation. The output of this procedure is a vector of size $C+1$ which is passed through a softmax to compute class probabilities. It should be noted that using a more sophisticated mechanism for scoring the model outputs against the prototypes is a potential extension of our work.
\subsection{Training labels}
The training labels for  each image are a sequence of (point, class) pairs organized in a specific order. The order is optional as long as it is consistent for all the images. We ordered the object labels from top-left to bottom right which is the order they would appear if we flattened the image, \textit{i.e.} lexicographical order by their $(y, x)$ coordinates. Finally, class indices are assigned randomly for each task (between 0 and C-1).

\subsection{Loss}
The loss function consists of two terms: a class-agnostic term which is responsible for the localization of objects by the attention module, and a classification term for classifying the localized objects.  
For the former we use the sum of KL-divergences between the generated attention maps for the objects in the query image and the corresponding Gaussian maps centered from the annotation points. For the latter, we use cross-entropy loss between the predicted class scores and the corresponding labels. For an efficient training, these two losses should be carefully weighted with respect to each other. At the beginning of the training the focus is more on teaching the model to distinguish the objects from the background. As the model gets better at sequentially attending to and localizing the objects, it becomes more important to focus on classification. To encourage this, we use an adaptive weighing scheme to gradually decrease the weight of the class-agnostic loss as the training proceeds. Our experiments proved the effectiveness of this strategy.
The total loss is given by:
 \begin{align}
    \mathcal{L} &= \lambda_1^t \sum_t \text{KL}(\alpha_t||G_t) +  \lambda_2^t \sum_t \text{CE}(y_t||c_t),
\end{align}
where KL stands for Kullback–Leibler divergence, CE is the cross-entropy loss, and with $t$ indexing the time-step, $\alpha_t$ and $G_t$ represent the generated attention mask for the query image and the Gaussian kernel from the labels respectively. $y_t$ is the one-hot class label and $\lambda_1^t$ and $\lambda_2^t$ are the time-varying weights associated to the KL and CE loss terms.

\section{Cafeteria: a diverse multi-object counting dataset} \label{cafe_ds}
Object detection datasets such as PASCAL VOC~\cite{Everingham10} and COCO~\cite{lin2014microsoft} have been the most popular datasets used for object counting. The main drawback of these datasets is that most images contain only a small number of instances from few categories. This simplifies the multi-class object counting task, specially in the few-shot context. Additionally, most of the object counting specific datasets consist of images with a high density of objects from only a single category: people~\cite{chan2008privacy, chen2012feature, zhang2016single, zou2019attend}, cars~\cite{de2015pklot, onoro2016towards}, penguins~\cite{arteta2016counting} or cells~\cite{marsden2018people, xie2018microscopy}.
However, in practical applications multi-class object counting tasks include more challenging scenarios where each image contains objects from several different classes. In order to address these challenges, more sophisticated datasets are required. As part of our effort to address this problem, we introduce the Cafeteria dataset, a diverse dataset with a suitable number of categories that can be used for few-shot object counting. 

\begin{figure}[t]
\center
    \includegraphics[width=0.63\textwidth]{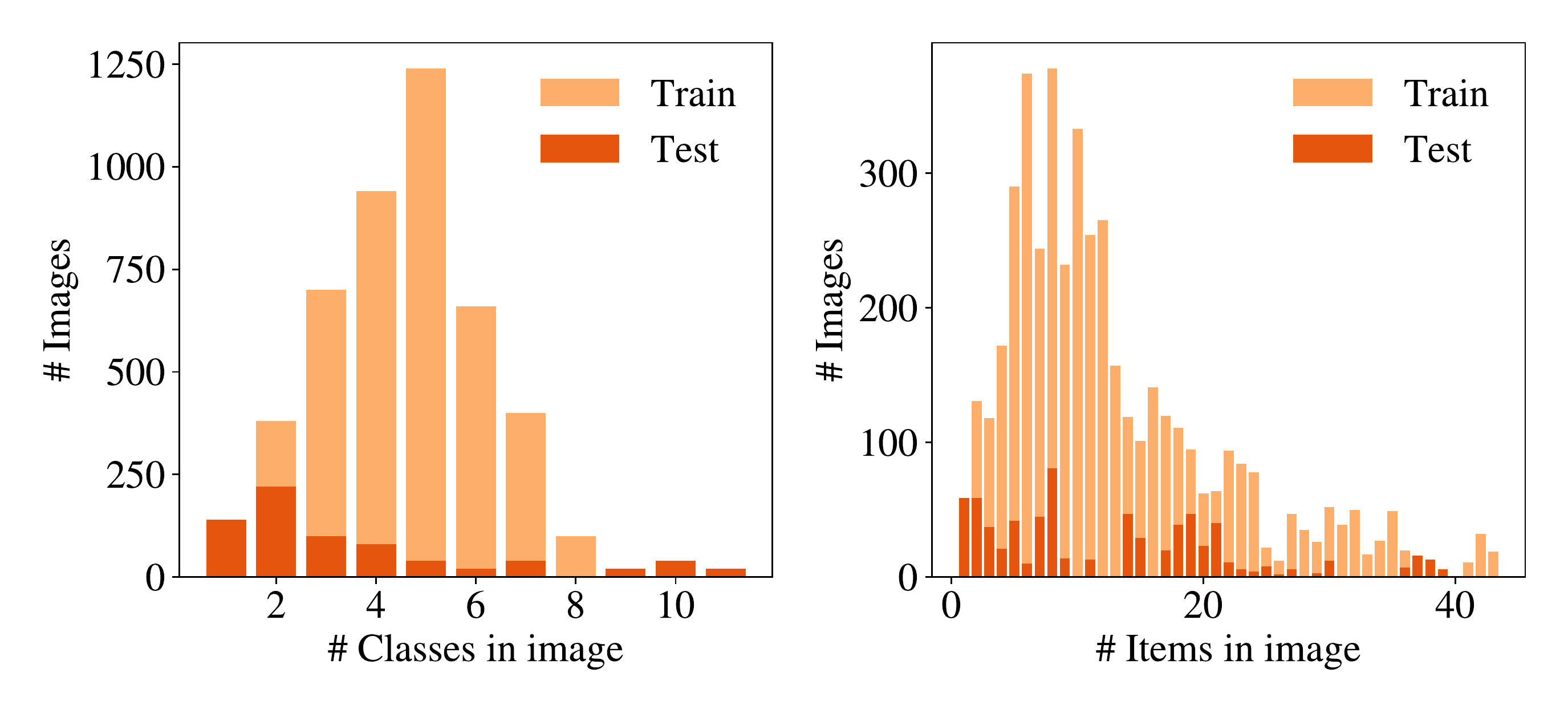}
    \caption{Distribution of the Cafeteria dataset.}
    \label{fig: cafeteria-stats}
\end{figure}

\begin{figure}[t]
\centering
    \includegraphics[width=.8\textwidth]{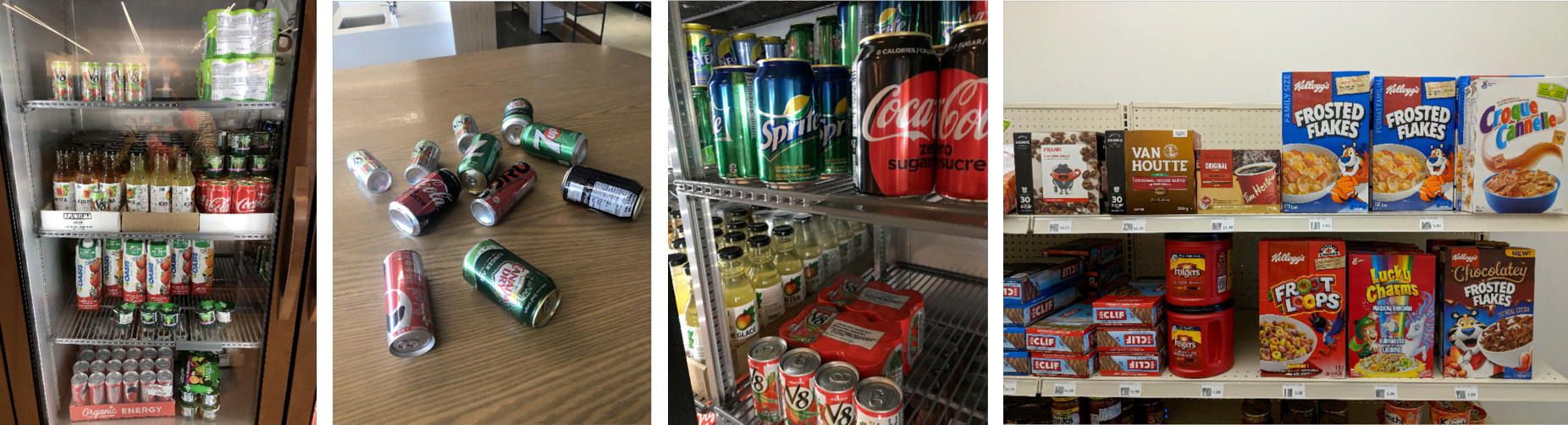}
    \caption{Examples of the Cafeteria dataset}
    \label{fig: cafeteria-examples}
\end{figure}

The Cafeteria dataset is composed of images of grocery items on shelves, fridges and other surfaces \cmmnt{(tables, floors, etc.)}taken with cellphone and security cameras (Figure \ref{fig: cafeteria-examples}). The images were labelled with point-level annotations by skilled annotators. This is a complex dataset where items appear in different shapes, sizes, colors and orientations with a wide variety of backgrounds. The dataset contains several images where objects are densely-packed, providing samples with stacking and occlusion. Unlike the supermarket dataset presented in~\cite{trax}, our images were taken from significantly different angles and distances, and the objects of the same class are not always grouped together. Moreover, our dataset exhibits a high variety in the number of classes and items per image, as can be seen in Table \ref{table: stats} and Figure \ref{fig: cafeteria-stats}.
\vspace{-5mm}

\begin{table}
\small
\centering
\caption{Statistics of the Cafeteria and FS-COCO datasets}
\resizebox{0.45\textwidth}{!}{
\begin{tabular}{@{}l|cc|cc@{}}
\toprule
&\multicolumn{2}{c}{Cafeteria dataset} & \multicolumn{2}{c}{FS-COCO dataset}\\
\toprule
 & Train & Test& Train & Test \\ \midrule
\# Images & 5244 & 901 & 7084 & 5465\\
\# Classes & 41 & 27 & 38 & 41\\
\# Tasks   & 4520 & 720 & 268 & 165 \\
Avg. Classes / Image & 4.61 & 3.62 & 2.1 &  1.8\\
Avg. Objects / Image & 12.98 & 12.65 & 4.2 &  3.7 \\ \bottomrule
\end{tabular}}
\label{table: stats}
\end{table}
\vspace{-1cm}

\section{Experiments}

In order to demonstrate the generalizability of the proposed model, we train it on three different datasets separately and evaluate each version of the model on the test-sets of all three datasets. We use four different metrics for evaluation: the mean absolute error (MAE) and root mean square error (RMSE), as well as recall and precision metrics. For our baseline comparison, we use GMNet ~\cite{lu2018class}, a class-agnostic object counting model that uses a general matching architecture. To the best of our knowledge, this is the only work in few-shot object counting available in the literature. Finally, we report the outcome of our ablation studies in Section \ref{ablation}.

\subsection{Training Details}\label{training_det}
We followed a simple process for compiling few-shot datasets from existing object counting/detection datasets by defining each unique subset of classes as a task. See Section \ref{problem_def} for more detailed definitions of tasks). In our experiments we set the number of query images in each task to $1$, while the number of support images was randomly chosen to be between $3$ and $5$.
We trained the models end-to-end using the Adam optimizer \cite{kingma2014adam} with an initial learning rate of $0.00004$. The learning rate was decreased by a factor of $0.5$ every time the validation error did not improve after $40$ epochs. Each epoch consists of $100$ episodes of training. Unless otherwise mentioned, we used a default standard deviation of $\sigma = 8$ for the Gaussian kernels used for weighing the features.

\subsection{Datasets}
The three datasets we use in our experiments are the Cafeteria dataset (Section \ref{cafe_ds}), FSOD, and MS COCO.

\textbf{MS COCO:} in order to evaluate the model on a dataset with a large variance in the appearances and scales of objects in natural scenes, we compiled a relatively small few-shot dataset from MS COCO by splitting the $80$ classes into train and test and compiling meta-train and meta-test samples from the train and test classes respectively. We only retained images corresponding to tasks with sufficient number of images with that composition. We refer to this compiled version of MS COCO as \textit{FS-COCO}. Table \ref{table: stats} shows some statistics on the resulting dataset.

\textbf{FSOD:} the Few-Shot Object Detection dataset (FSOD) was introduced in~\cite{fan2019few} as a diverse dataset designed specifically for few-shot object detection. It consists of overall $1000$ categories with a $800/200$ split between train and test, and a total of $66000$ images with $182000$ bounding boxes. To adapt this diverse and challenging dataset to train our weakly-supervised approach, we took the center of the bounding boxes as the annotation points.

\subsection{Evaluation Strategy} 
Our evaluation method can be described as \textit{$C$-way $S$-shot} where $C$, the number of distinct categories in the image, varies across the tasks. We report the results as a function of $S$, the number of shots. Following the fewshot learning principle, we evaluate our models on classes unseen during training. The only exception is the model trained and tested on Cafeteria, for which the train and test classes overlap (the tasks are still new). The results in this case are still interesting as they show the object counting capability of the model on a challenging dataset. We use the MAE, RMSE, Recall and Precision metrics for evaluation. \footnote{More details on evaluation metrics are explained in supplementary material}

\subsection{Results}
The results of our model trained on Cafeteria, FSOD and FS-COCO datasets on all of the test-sets are shown in table \ref{table: train-on-all}. We can see that the proposed method outperforms the benchmark model in all nine scenarios, as indicated by all the metrics. Table~\ref{table: train-on-all} shows that training on Cafeteria results in a recall of $90\%$ and a precision of $59\%$ when tested on FSOD. Given the considerable difference between the distribution of the classes in these two datasets, the results highlight the generalizability of the model in dealing with truly different classes, \textit{i.e.} the true ``few-shot'' nature of how the model learns.

It is interesting to note that the results for the models trained on Cafeteria show the largest gap between our model and the benchmark. The reason could be that the Cafeteria dataset has much more classes per image on average than other datasts (see Table\ref{table: stats}), which makes the problem much more challenging. This suggests that our model is better equipped to learn on images with larger number of categories.
It is also noteworthy that the model trained on FS-COCO performs better on FSOD than on FS-COCO itself, which may be due to the challenging nature MS COCO.
In order to understand why the results are generally better on the FSOD test-set across the board, we have to remember that apart from the fact that the number of items per image are smaller for FSOD ($2.8$ items and $1.15$ distinct classes on average), the point-annotations on FSOD are taken from the bounding boxes, resulting in cleaner labels closer to the center of objects.
\vspace{-5mm}

\begin{table}
\center
\caption{Performance of models trained on one dataset and tested on another one with the corespondent number of shots ($S$)}
\small
\resizebox{\textwidth}{!}{
{\begin{tabular}{@{}l|l|cccc|cccc|cccc@{}}
\toprule
\multicolumn{2}{l}{Trained on} & \multicolumn{4}{c}{Cafeteria-train, $S=3$} & \multicolumn{4}{c}{FSOD-train, $S=5$} & \multicolumn{4}{c}{FS-COCO-train, $S=5$} \\
\toprule

Method & Tested on & MAE & RMSE & Recall & Precision& MAE & RMSE & Recall & Precision& MAE & RMSE & Recall & Precision \\ \midrule
\multirow{3}{*}{GMNet} & FS-Cafeteria & 14.15 & 16.19 & 0.23 & 0.78 & 3.32 & 3.76 & 0.51 & 0.75 & 2.54 & 2.74 & \textbf{0.74} & 0.16 \\
 & FS-COCO & 4.12 & 4.49 & 0.41 & 0.48 & 2.82 & 3.01 & 0.63 & 0.45 & 1.92 & 2.04 & 0.51 & 0.87 \\
 & FSOD & 6.65 & 6.69 & 0.36 & \textbf{0.66} & 2.33  & 2.35 & 0.65 & 0.67 & 2.10 & 2.12 & 0.78 & 0.17\\ \midrule
\multirow{3}{*}{Ours} & FS-Cafeteria & \textbf{1.53} & \textbf{1.95} & \textbf{0.79} & \textbf{0.78} & \textbf{1.58} & \textbf{1.78} & \textbf{0.57} & \textbf{0.92} & \textbf{1.92} & \textbf{2.15} & 0.43 & \textbf{0.94}\\
 & FS-COCO & \textbf{2.83} & \textbf{3.48} & \textbf{0.50} & \textbf{0.50} & \textbf{1.65} & \textbf{1.84} & \textbf{0.63} & \textbf{0.79} & \textbf{1.61} & \textbf{1.79} & \textbf{0.59} & \textbf{0.88}\\
 & FSOD & \textbf{2.93} & \textbf{2.95} & \textbf{0.90} & 0.59 & \textbf{0.86} & \textbf{0.88} & \textbf{0.76} & \textbf{0.99} & \textbf{1.99} & \textbf{2.02} & \textbf{0.92} & \textbf{0.83}\\ \bottomrule
\end{tabular}}}
\label{table: train-on-all}
\end{table}

\vspace{-8mm}

\subsubsection{Comparison with Few-shot detection approaches}
In table \ref{table:fsod} we compare our results with the state-of-the-art few-shot object detection approaches \cite{fan2019few, kang2019few, chen2018lstd}. It is important to keep in mind that unlike our approach, the detection-based approaches require bounding boxes for training. Moreover, in order to convert detection-based outputs to counting results we counted every detected object with a higher IoU than 0.5.

\begin{table}[]
\small
\center
\caption{Performance comparison with few-shot detection-based approches, $S=5$}

\resizebox{0.37\textwidth}{!}
{\begin{tabular}{@{}l|lcc@{}}
\toprule
 Dataset & Method & Recall & Precision \\  \midrule
 \multirow{4}{*}{FS-COCO} & FSOD~\cite{fan2019few} & 0.51 & 0.41 \\
 & FODvFR ~\cite{kang2019few} & 0.28 & 0.12 \\
 & LSTD(YOLO)~\cite{chen2018lstd} & 0.20 & 0.09 \\
 & Ours & \textbf{0.59} & \textbf{0.88} \\
 \hline
 \multirow{2}{*}{FSOD} & FSOD~\cite{fan2019few} & 0.49 & 0.44 \\
 & Ours & \textbf{0.76} & \textbf{0.99} \\ \bottomrule
\end{tabular}}
\label{table:fsod}
\end{table}

This comparison results show that even though detection approaches rely on bounding boxes in training, they are considerably outperformed by our approach.

\subsection{Ablation Studies} \label{ablation}
\cmmnt{For the ablation studies, we experimented with changing or removing some components from our model as explained below.}
\textbf{Relying on attention for support images:} one might wonder what would happen if extracting features for the class prototypes relies on the attention module instead of the Gaussian maps generated from the labels, similar to what happens with query images. To find out, we trained a version of the model without the label Gaussian maps for the support images. In this case, the two folds of the model in Figure \ref{fig: model} are exactly the same. Table \ref{table: ec-guide} shows the results of this experiment against the original model. Both models are trained and tested on FSOD.
The significant drop in the precision for the model that does not use the Gaussian kernels implies that the module is taking full advantage of them. \\

\textbf{Removing the encoded coordinates:} concatenating the encoded coordinates (EC) at the end of the extracted feature maps from the image makes the model more location aware. This is specifically helpful in following the predetermined order of objects in the image. Table \ref{table: ec-guide} compares the performance of the model trained on FSOD train-set and tested on FSOD test-set with and without the encoded coordinates, indicating that including them improves the performance of the model considerably, specially in terms of the recall rate.

\vspace{-5mm}

\begin{table}
\center
\small
\caption{Effect of the encoded coordinates and guiding the model when extracting the features from the support set}

\resizebox{0.43\textwidth}{!}
{\begin{tabular}{@{}rrrrr@{}}
\toprule
 & MAE & RMSE & Recall & Precision \\ \midrule
Full approach & \textbf{0.86} & \textbf{0.88} & 0.76 & \textbf{0.99} \\
Without EC & 0.91 & 0.93 & 0.71 & 0.96  \\
Without guide & 6.25 & 6.36 & \textbf{0.93} & 0.51  \\ \bottomrule
\end{tabular}}
\label{table: ec-guide}
\end{table}


\textbf{Altering the standard deviation of Gaussian kernels:} for very low density datasets such as FSOD and COCO, estimating the standard deviation of the Gaussian kernel from the point level annotations using the approach suggested in \cite{zhang2016single} is very inaccurate if not impossible since the number of objects per image is usually very small. Therefore, for these datasets we use a fixed default value for the standard deviation. Table \ref{table: sigma} shows the results of varying $\sigma$. All the models in this table have been trained and tested on FSOD.

It can be seen that the optimum value of $\sigma$ for the FSOD dataset is $8$. The optimal value varies for each dataset, depending on the scale of the objects and their composition. A potential expansion of this work, could be enforcing the model to predict the value of $\sigma$ for each class in each image by adding an extra regression head.

\vspace{-5mm}

\begin{table}
\small
\center
\caption{Effect of the standard deviation of the Gaussian kernels}
\resizebox{0.30\textwidth}{!}{
\begin{tabular}{@{}rrrrr@{}}
\toprule
$\sigma$ & MAE & RMSE & Recall & Precision \\ \midrule
 4 & 1.83 & 1.85 & \textbf{0.91} & 0.64 \\ 
 6 & 1.74 & 1.77 & \textbf{0.91} & 0.65 \\
 8 & \textbf{0.86} & \textbf{0.88} & 0.76 & \textbf{0.99} \\ 
 10 & 0.91 & 0.93 & 0.79 & 0.93 \\ \bottomrule
\end{tabular}}

\label{table: sigma}
\end{table}

\begin{figure}[t]
    \centering
    \includegraphics[width=0.55\textwidth]{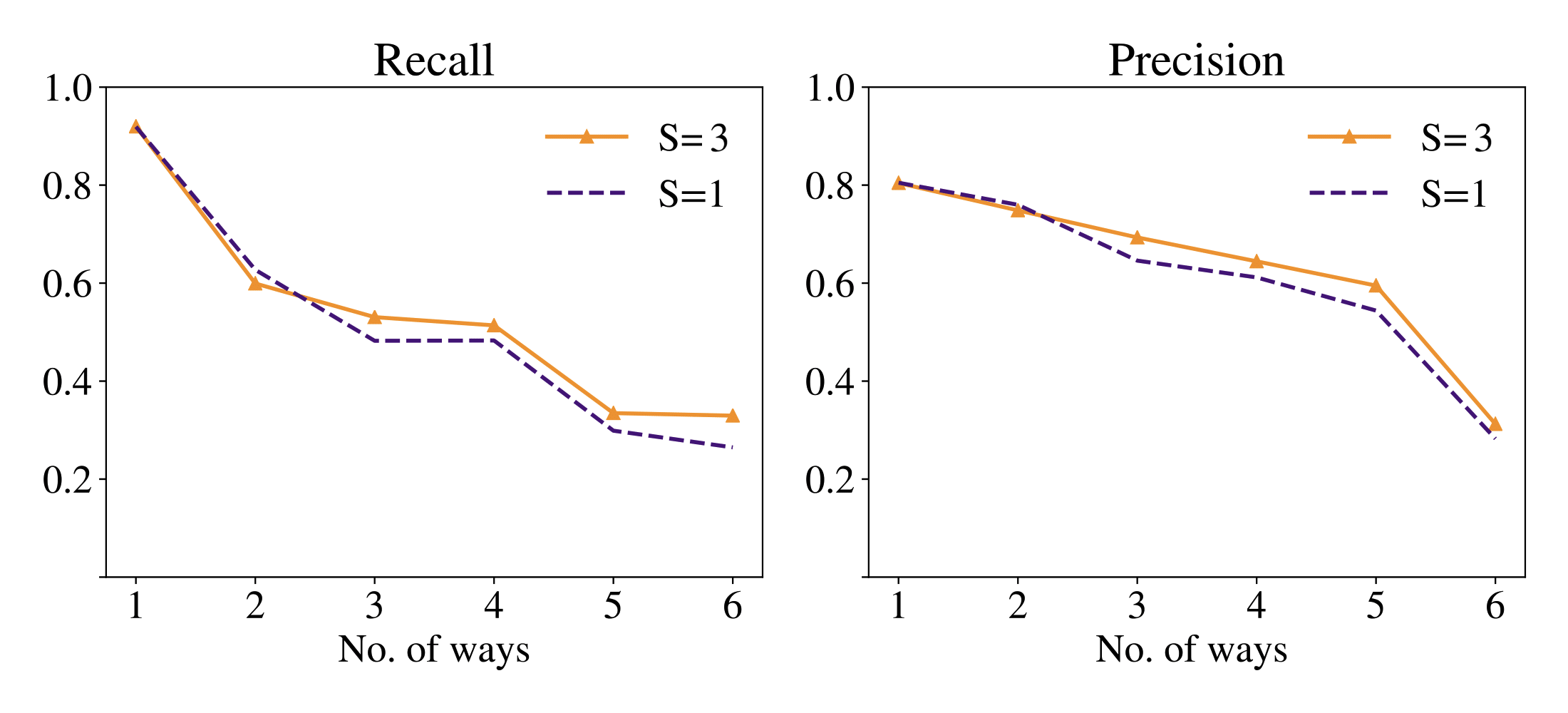}
    \caption{Performance as a function of the number of shots $S$ and ways $C$. The model was trained and evaluated on Cafeteria.}
    \label{fig: shots}
\end{figure}

\vspace{10mm}
\textbf{Number of shots:} in Figure \ref{fig: shots}, we observe that increasing the number of shots slightly improves the performance of the model, specially when the number of ways are larger. The reason is that as the number of support images increases, the prototype vectors are averaged over larger number of objects, hence smoothing out the noise in features. We plan to study other approaches for aggregating multiple features from multiple instances of the same class in the support set.

\section{Conclusion}
\cmmnt{In this work, we addressed the problem of point-level few-shot multi-class object counting.

We proposed a loss function that consists of two terms, i) a class agnostic attention term that sequentially attends to objects in an image and ii) a classification term that correctly classifies the localized objects. Our attention mechanism is employed on an adapted prototypical-based few-shot approach, that enables the model to learn localizing objects with only a very few samples from given classes.
We are state-of-the-art on three challenging datasets: i) MS- COCO ii) FSOD iii) Cafeteria. Cafeteria dataset is a dataset that we particularly designed for the task of few-shot object counting. Cafeteria dataset has a different distribution with natural images. We also presented our results when we train the model on one dataset and test it on totally different data distribution.}

We addressed the little-studied problem of few-shot multi-class object counting with point-level supervision. We presented results on three challenging datasets with diverse distributions, namely, FSOD, a few-shot subset of MS COCO, and Cafeteria; a dataset we introduced specifically for object counting. Our model employs an intuitive approach to object counting where the objects are first sequentially localized through a class-agnostic attention mechanism before being classified via a prototypical-based few-shot scheme. Our loss function, crafted to reflect this two-step approach, combines two terms corresponding to the localization and classification steps. \cmmnt{ with a time-varying weighting scheme.}

Our approach mitigates the challenge of multi-class object counting by reducing the dependency on labelled data, at the same time simplifying the labelling process by relying on point-annotations only. We believe that our work motivates further research into a problem that is of increasing interest in many applications, namely, few-shot multi-class object counting. One such application is inventory tracking, where data-scarcity due to the rapidly changing composition of items, as well as the complexity of the problem setup, poses a real challenge to existing approaches.

\clearpage
%
%
\bibliographystyle{splncs04}

\newpage

\end{document}